\documentclass[11pt]{article}

\usepackage{acl2015}
\usepackage{times}
\usepackage{url}
\usepackage{latexsym}
\usepackage{graphicx}
\usepackage{todonotes}

\graphicspath{ {crowd/img/} }
\title{Presenting a New Dataset for the Timeline Generation Problem}

\author{Xavier Holt \\
  University of Sydney \\
  {\tt xhol4115@}\\
  {\tt uni.sydney.edu.au}\\\And
  Will Radford \\
  Hugo.ai \\
  {\tt wradford@}\\
  {\tt hugo.ai} \\\And
  Ben Hachey \\
  University of Sydney \\
  {\tt ben.hachey@}\\
  {\tt sydney.edu.au} \\}

\date{}

\begin{document}
\maketitle
\begin{abstract}
	The timeline generation task summarises an entity's biography by selecting stories representing key events from a large pool of relevant documents.
%Timeline generation is a summarisation method for distilling vast amounts of entity-related information into a concise and readable format. 
This paper addresses the lack of a standard dataset and evaluative methodology for the problem.

We present and make publicly available a new dataset of 18,793 news articles covering 39 entities. 
For each entity, we provide a gold standard timeline and a set of entity-related articles. 
We propose ROUGE as an evaluation metric and validate our dataset by showing that top Google results outperform straw-man baselines.
\end{abstract}

%%%% Sections %%%%
%%INTRO
\section{Introduction}
Information is more readily available in greater quantities than ever before. 
Timeline generation is a recent method for summarising data -- taking a large pool of entity-related documents as input and selecting a small set that best describe the key events in the entity's life.
%One recent method for summarising massive amounts of data and presenting it in an accessible way is timeline generation.
%Timeline generation is a method of summarising large sets of entity-related documents. 
%By some measure it aims to find the $k$ best documents for summarising the corpus.
There are several challenges in evaluation: (1) finding gold-standard timelines, (2) finding corpora from which to draw documents to build timelines, and (3) evaluating system timelines.

Standard practice for the first challenge is to make use of existing timelines produced by news agencies \cite{Chieu:2004id,Yan2011}, but these are constrained by the tight editorial focus on prominent entities and depends on well-funded news agencies.
Another approach is to annotate new timelines from the web for domains of choice.
\newcite{Wang2013} do this, but do not make their data available for direct comparison.
%their annotation protocol such that it can be easily replicated for new domains.
Regarding the second challenge, access to the document pool used during the annotation process is also important, as any system must have a reasonable set from which to choose.

Previous approaches have used drawn on working in document summarisation, using ROUGE \cite{lin2004rouge} to evaluate timeline generation
%Timelines can be approximately compared to gold-standard using the ROUGE metric \cite{lin2004rouge} and this is common in the field 
\cite{Chieu:2004id,Yan2011,Yan2011a,Ahmed:2012vh,Wang2013}.
This is convenient as
%This is especially important when working with noisy and redundant web corpora, as 
each element in a timeline can represent a \emph{story} which can be equivalently described by many different documents.
However, previous work has not validated the use of ROUGE for evaluating timeline generation.

We present a general framework for creating a crowd-sourced datasets for evaluating timeline generation, including
%We present a broad, rigorous method for 
choosing a set of entities, deriving articles for annotation from Wikipedia,
%selecting a minimally sufficient set of articles, 
and annotating these articles to generate a gold standard. 

The dataset covers a broad range of entities with different levels of news-coverage or publicity. We provide gold-standard timelines for each entity, as well as a larger pool of topically-linked documents for further development.
We analyse the dataset, showing some interesting artifacts of crowd-worker importance judgements and use ROUGE evaluation to verify that the crowd-annotations correlate well with Google News\footnote{\url{https://news.google.com}} rankings.
This reflects favourably on Google News, suggesting that it is a strong baseline for timeline generation.
We release the dataset generated through this process in the hope that it will be useful in providing common benchmarks for future work on the timeline generation task.\footnote{\url{https://github.com/xavi-ai/tlg-dataset}}
%the proposed method and dataset provide new resources for researchers working in timeline generation.

\label{sec:intro}

%%DSET METHODOLOGY
\section{Data Collection}

We now detail how we choose entities and collect a corpus for annotating gold-standard timelines and evaluating models.
We have taken care to design a general experimental protocol that can be used to generate entities from a range of domains.

We begin by choosing a domain (politics) and two regions (The USA/Australia). We motivate this choice of domain by noting its large media interest, polarising entities and diverse range of topics.
We choose several entities from each region a priori -- 39 in total. The rest of our entity-set is then generated through a process of bootstrapping. 
At each iteration, we use our current entities as seeds. For each seed-entity, we performed a Google News query. An entity name was defined as either the title of the entity's Wikipedia page or \{\#Firstname \#Lastname\} if they did not have a Wiki.
We choose five articles from the first page of results. For each article we manually identify all other previously unseen entities and add them to our set. We continue this process of bootstrapping, using our newly included entities as seeds for the next iteration. Once we have a sufficient number of entities, we terminate the process. 

This process can be viewed as \emph{bootstrap sampling}, weighted by the probability an entity occurs in one of the articles we retrieve. By doing so we  provide a realistic set of entities with varying levels of popularity and coverage. In Section~\ref{sec:analysis} we show that this process results in a wide distribution of mentions and reference-timeline sizes.

As well as relevant entities, we also provide a corpus of relevant documents from which we can construct their timelines. 
Each document in the corpus includes URL, publishing date and other metadata.
These were obtained by performing Google News queries on our entities, and retrieving the resulting URLs. 
As our timelines should cover a wide range of time, we set the time-range on the query to `archives'. 
This has the effect of returning articles from a broader period of time, mitigating the default recency bias.

In total, there are 15,596 articles. The minimum, median and maximum number of articles per entity was 54, 464 and 985 respectively. 
By including this corpus with our gold-standards we aim to provide a complete dataset for the timeline generation task.
\label{sec:method}

%%ANNOTATION
\section{Data Annotation and Gold-Standards}

We present a general framework for formulating gold-standard timeline generation as an annotation task. This involves two components -- using Wikipedia to generate a minimal set of sufficient links, and the formulation of the problem as an annotation task.

\subsection{Article Selection}
\label{ssec:cand}

Annotation is cost-sensitive to the size of the task. As such, attempting to annotate the whole corpus of over 15,000 articles is infeasible. We propose a method to reduce the size of our task while maintaining the quality of the underlying timeline.
For our article selection process, we need to fulfil the following criteria:

\begin{itemize}
\item {Coverage}: Our set of articles should have good coverage. Timelines should cover a broad range of time-periods and events. As such, the dataset we derive our reference timelines from must also share this property.
\item {Manageability}: Each entity-article pair will be subject to a number of crowd-judgments. As such, it's important to balance coverage with total data-set size. 
\item {Informativeness}: Ideally we desire the articles to be of a high quality.
\end{itemize}

To meet these criteria, we scrape the external (non-Wiki) links from an entity's Wikipedia page. We motivate this decision by first noting the Wikipedia guidelines on verifiability\footnote{https://en.wikipedia.org/wiki/Wikipedia:Verifiability}:

\begin{quotation}
	Attribute all quotations and any material challenged or likely to be challenged to a reliable, published source using an inline citation. The cited source must clearly support the material as presented in the article.
\end{quotation}

These standards of verifiability are not universally followed. Nevertheless, where they are we expect reasonable entity coverage and informativeness.
After removing invalid URLS, we identify 3,197 articles for annotation.

\subsection{Crowd-task Formulation}
\label{ssec:crowd}

We formulate timeline generation as an annotation task by reducing it to a simple classification problem.
A single judgment is on the level of an entity-article pair. An annotator is given the first paragraph of, and a link to, the entity's Wikipedia page. They then follow a given link, and perform a two-stage classification task.

The annotators first determine whether a link is valid. A valid article is one that covers a single event in the target entity's life.
They then indicate the importance of an article's when considering the story of the entity. There was a choice of three labels:

\begin{itemize}
    \item{Very important}: key events which would be included in a one-page summary or brief of the entity. 
    \item{Somewhat important}: newsworthy events that might make it into a broader Biography, but not of critical relevance. 
    \item{Not important}: events which are mundane or unimportant.
\end{itemize}

For our annotations, we use the CrowdFlower\footnote{https://www.crowdflower.com} platform.
On average, three judgments by a trusted user were made per row. A trusted user was one whose annotations agreed with our expert's across a validation set ($n=48$) at least 80\% of the time. This was then aggregated into a classification label.
In addition, CrowdFlower also provides a confidence measure on each judgment -- a score of agreement, weighted by trust of the crowd-worker.
Gold standard timelines comprise the articles that are judged to be both `valid' and `very important'. There were 2,601 `valid articles' and 217 `very important'.

\label{sec:annota}

%%ANALYSIS
\section{Analysis of Gold-Standards}
\label{sec:ana}

\begin{figure}[t]
  \includegraphics[width=\columnwidth]{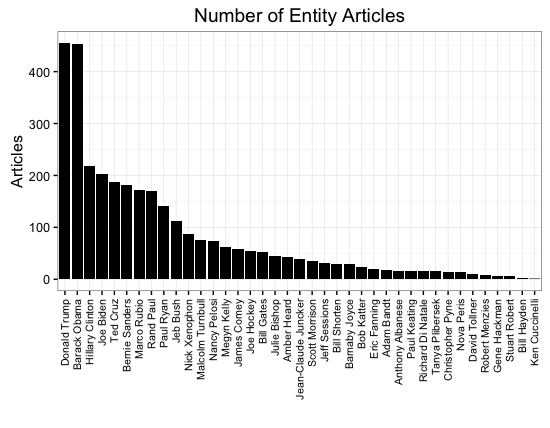}
  \caption{Number of different articles by entity. A small number of entities are responsible for a large number of articles.}
  \label{fig:nArt}
\end{figure}

We see that particularly prominent entities are responsible for a large portion of the articles. `Barack Obama' and `Donald Trump' each have over four hundred articles each. In fact, the six most prominent entities account for over half of all total articles (Figure ~\ref{fig:nArt}).

\paragraph{Very Important Articles} The `very important' articles make up our gold-standard timelines. The mean and median number of articles per entity is 5.56 and 2 respectively.

\begin{figure}[t]
  \includegraphics[width=\columnwidth]{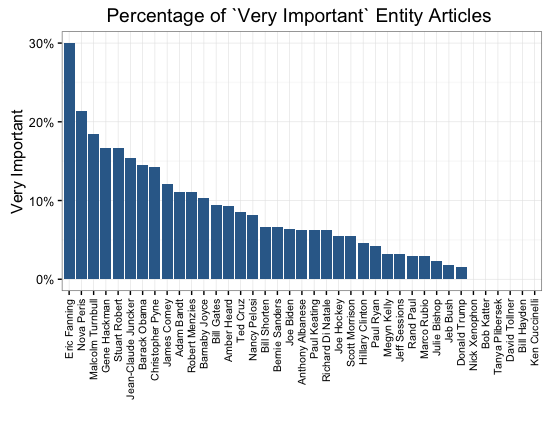}
  \caption{Percentage of very important articles by entity.}
  \label{fig:vImpP}
\end{figure}

There are some interesting properties that emerge. `Barack Obama' and `Donald Trump' each have around the same number of articles. The former has 14.6\% articles deemed `very important' -- the latter only 1.5\% (Figure~\ref{fig:vImpP}). It is a given that certain entity's will be involved in more newsworthy events than others. However, to have such a large\footnote{Or tremendous?} discrepancy -- considering to that all articles were deemed necessary to reference in an entity's Wiki -- is curious. We believe the proportion of `very important' articles for a given entity is an interesting avenue of future research.

\begin{figure}[t]
  \includegraphics[width=\columnwidth]{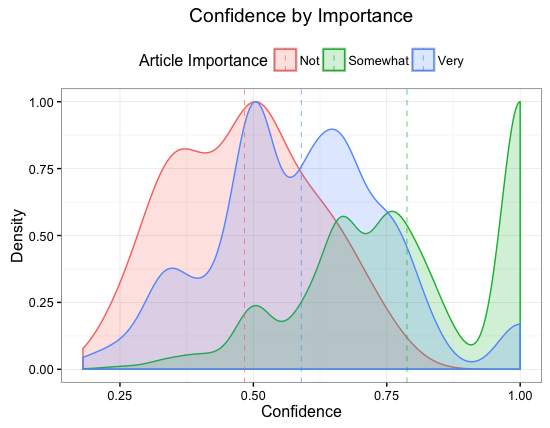}
  \caption{Distribution of confidence for three different levels of article importance. Mean-values are indicated by a dashed line.}
  \label{fig:conf}
\end{figure}

\begin{figure}[t]
  \includegraphics[width=\columnwidth]{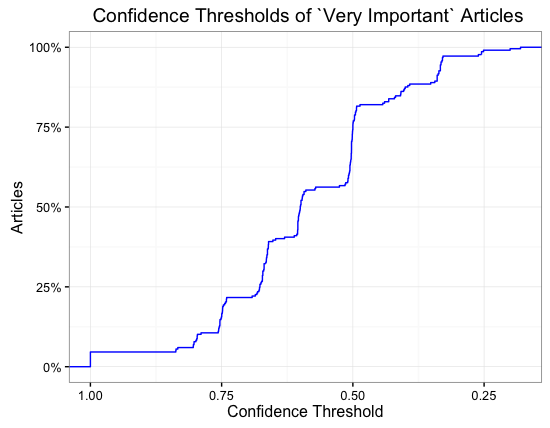}
  \caption{Percentage of `very important' articles meeting a given confidence threshold.}
  \label{fig:ecdf}
\end{figure}

\paragraph{Confidence}  `Very important' articles have a mean confidence of 0.60. Only 4.6\% of articles received a unanimous 1.00 confidence score (Figure~\ref{fig:conf}). However, three-quarter of the `very important articles' had a confidence over 0.50 (Figure~\ref{fig:ecdf}).
`Somewhat important' articles have the highest overall confidence, with a mean value of 0.76. Over a third of these articles had a confidence score of 1.00 (Figure~\ref{fig:conf}). This is somewhat understandable. Intuitively, `somewhat important' is the default prior -- we would expect most articles to fall in this category. 

\label{sec:analysis}

%%EVALUATION
\section{Evaluation}
%\label{sec:eval}

For our evaluation pipeline, we adopt the approach of a number of papers in the field \cite{Wang2013,Yan2011,Yan2011a} in using the ROUGE metric \cite{lin2004rouge}. ROUGE was first used in automatic summarisation evaluation. It is similar to the BLEU measure for machine translation \cite{BLEU}.
In terms of timeline evaluation, quality is measured by the amount of overlapping units (e.g. word n-grams) between articles in a system timeline and articles in a reference timeline. For details on how ROUGE scores are calculated, please refer to the original paper \cite{lin2004rouge}.
For our purposes, articles annotated as `valid' and `very important' are taken to be components of an entity's reference timeline. We use the ROUGE-F measure over unigrams and bigrams ($n = 1, 2$).
\label{sec:eval}

%%BASELINE
\section{Benchmarks and System Validation}

In this section we use our supplemental dataset of articles generated by Google News to validate and benchmark the task.

\begin{figure}[t]
  \includegraphics[width=\columnwidth]{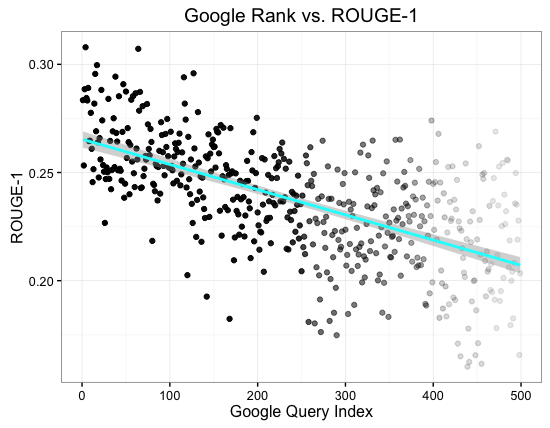}
	\includegraphics[width=\columnwidth]{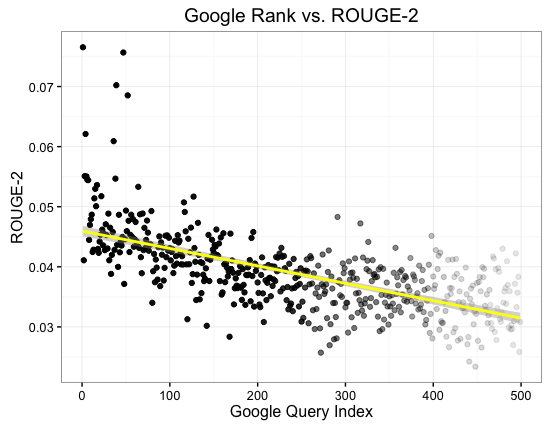}
  \caption{Google search rank for a given article vs. ROUGE score. Each point is an average across entities. The intensity of the point is a measure of confidence}
  \label{fig:rou_one}
\end{figure}

\paragraph{ROUGE vs. Search Rank} For a given news query, an article's rank is a signal of its important and centrality. It is reasonable to expect then that the better an article's search-rank, the more likely it is to appear in an entity's timeline.
This appears to be the case. For both the ROUGE-1 and -2 measures, there is a clear negative correlation between an article's average score and index (Figure~\ref{fig:rou_one}).

\paragraph{Benchmarks}

\begin{figure}[t]
  \includegraphics[width=\columnwidth]{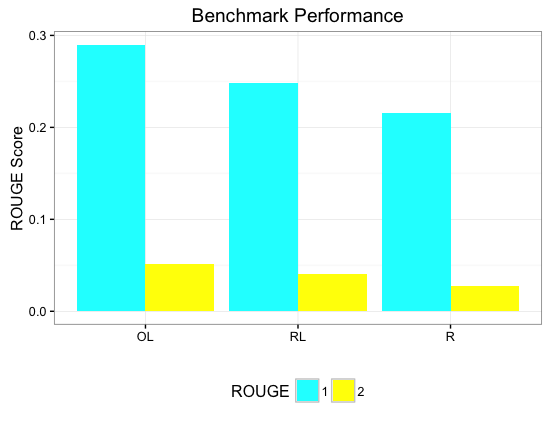}
  \caption{Benchmark performance. As expected, OL outperforms RL which in turn outperforms R for both ROUGE-1 and -2.}
  \label{fig:bench}
\end{figure}

For a given entity timeline, we include the following three benchmarks -- Random (R): 15 articles are sampled from the entire corpus. Random+Linked (RL): 15 articles linked to the entity are sampled. Ordered+Linked (OL): the 15 highest ranked articles for an entity are chosen.
Reassuringly, we see that OL outperforms RL which outperforms R for both ROUGE-1 and ROUGE-2 scores (Figure~\ref{fig:bench}).
OL received scores of 0.290 (ROUGE-1) and 0.051 (ROUGE-2) (Table~\ref{my-label}). This can be taken as a strong benchmark for future timeline generation models trained and evaluated using this dataset.

\begin{table}[t]
\centering
\begin{tabular}{lcc}
\hline
\textbf{} & \multicolumn{1}{l}{\textbf{ROUGE-1}} & \multicolumn{1}{l}{\textbf{ROUGE-2}} \\ \hline
OL        & 0.290                                & 0.051                                \\
RL        & 0.248                                & 0.041                                \\
R         & 0.2052                               & 0.027                                \\ \hline

\end{tabular}

\caption{F-Scores for Benchmark Systems}
\label{my-label}
\end{table}

\label{sec:baseline}

%%CONCLUSION
\section{Conclusion and Future Work}

In this paper we have developed, analysed and justified a new dataset for the timeline generation problem. 
There are several interesting avenues for future work. 
The most obvious is the development of new timeline-generation systems using this dataset. 
There are also still problems to be solved with the process of evaluating timeline models, but we hope that the framework described above allow researchers to easily generate evaluation datasets for timeline generation. 

\label{sec:concl}

\bibliographystyle{acl}
\bibliography{crowd/biblio.bib}

\begin{thebibliography}{}

\bibitem[\protect\citename{Ahmed and Xing}2012]{Ahmed:2012vh}
Amr Ahmed and Eric~P Xing.
\newblock 2012.
\newblock {Timeline: A Dynamic Hierarchical Dirichlet Process Model for
  Recovering Birth/Death and Evolution of Topics in Text Stream}.
\newblock {\em CoRR abs/1203.3463}.

\bibitem[\protect\citename{Chieu and Lee}2004]{Chieu:2004id}
Hai~Leong Chieu and Yoong~Keok Lee.
\newblock 2004.
\newblock {\em {Query based event extraction along a timeline}}.
\newblock ACM, New York, New York, USA, July.

\bibitem[\protect\citename{Lin}2004]{lin2004rouge}
Chin-Yew Lin.
\newblock 2004.
\newblock Rouge: A package for automatic evaluation of summaries.
\newblock In Stan~Szpakowicz Marie-Francine~Moens, editor, {\em Text
  Summarization Branches Out: Proceedings of the ACL-04 Workshop}, pages
  74--81, Barcelona, Spain, July. Association for Computational Linguistics.

\bibitem[\protect\citename{Papineni \bgroup et al.\egroup }2002]{BLEU}
Kishore Papineni, Salim Roukos, Todd Ward, and Wei-Jing Zhu.
\newblock 2002.
\newblock Bleu: a method for automatic evaluation of machine translation.
\newblock In {\em Proceedings of 40th Annual Meeting of the Association for
  Computational Linguistics}, pages 311--318, Philadelphia, Pennsylvania, USA,
  July. Association for Computational Linguistics.

\bibitem[\protect\citename{Wang}2013]{Wang2013}
Tao Wang.
\newblock 2013.
\newblock {Time-dependent Hierarchical Dirichlet Model for Timeline
  Generation}.
\newblock {\em arXiv preprint arXiv:1312.2244}.

\bibitem[\protect\citename{Yan \bgroup et al.\egroup }2011a]{Yan2011}
Rui Yan, Liang Kong, Congrui Huang, Xiaojun Wan, Xiaoming Li, and Yan Zhang.
\newblock 2011a.
\newblock {Timeline Generation through Evolutionary Trans-Temporal
  Summarization.}
\newblock {\em EMNLP}, pages 433--443.

\bibitem[\protect\citename{Yan \bgroup et al.\egroup }2011b]{Yan2011a}
Rui Yan, Xiaojun Wan, Jahna Otterbacher, Liang Kong, Xiaoming Li, and Yan
  Zhang.
\newblock 2011b.
\newblock {Evolutionary timeline summarization}.
\newblock In {\em the 34th international ACM SIGIR conference}, page 745, New
  York, New York, USA. ACM Press.

\end{thebibliography}

\end{document}